\documentclass[letterpaper, 10 pt, conference]{ieeeconf}  

\usepackage{amsmath}
\usepackage{amssymb}
\usepackage{multirow}
\usepackage{booktabs}
\usepackage{float}
\usepackage{makecell}
\usepackage{mathtools}
\usepackage{ragged2e}
\usepackage{epsfig}
\usepackage{graphicx}
\usepackage{subcaption}
\usepackage{pifont}

\IEEEoverridecommandlockouts                              

\title{\LARGE \bf
Sparse Laneformer
}

\author{Ji Liu$^{*}$, Zifeng Zhang$^{*}$, Mingjie Lu, Hongyang Wei, Dong Li, Yile Xie, Jinzhang Peng, \\
Lu Tian, Ashish Sirasao, Emad Barsoum 
\thanks{$^{*}$Equally contributed to the work.}
\thanks{Ji Liu, Zifeng Zhang, Mingjie Lu, Hongyang Wei, Dong Li, Yile Xie, Jinzhang Peng, Lu Tian, Ashish Sirasao and Emad Barsoum are all with AEAI Group, Advanced Micro Devices, USA}
}

\begin{document}

\maketitle
\thispagestyle{empty}
\pagestyle{empty}

\begin{abstract}

  Lane detection is a fundamental task in autonomous driving, and has achieved great progress as deep learning emerges. Previous anchor-based methods often design dense anchors, which highly depend on the training dataset and remain fixed during inference. We analyze that dense anchors are not necessary for lane detection, and propose a transformer-based lane detection framework based on a sparse anchor mechanism. To this end, we generate sparse anchors with position-aware lane queries and angle queries instead of traditional explicit anchors. We adopt Horizontal Perceptual Attention (HPA) to aggregate the lane features along the horizontal direction, and adopt Lane-Angle Cross Attention (LACA) to perform interactions between lane queries and angle queries. We also propose Lane Perceptual Attention (LPA) based on deformable cross attention to further refine the lane predictions. Our method, named Sparse Laneformer, is easy-to-implement and end-to-end trainable. Extensive experiments demonstrate that Sparse Laneformer performs favorably against the state-of-the-art methods, e.g., surpassing Laneformer by 3.0\% F1 score and O2SFormer by 0.7\% F1 score with fewer MACs on CULane with the same ResNet-34 backbone.

\end{abstract}

\section{INTRODUCTION}

Lane detection is a fundamental task in autonomous driving, which predicts lane location in a given image. It plays a critical role in providing precise perception of road lanes for the advanced driver assistant system or autonomous driving system. 

Before deep learning methods emerged in lane detection, traditional computer vision methods were used (e.g., Hough lines~\cite{berriel2017ego,assidiq2008real}) but suffered from unsatisfactory accuracy. Due to the effective feature representations of convolutional neural networks (CNN), CNN-based methods~\cite{qu2021focus,wang2022keypoint,tabelini2021cvpr,8624563} have achieved remarkable performance and exceed traditional methods largely. These CNN-based methods can be mainly divided into segmentation-based, parameter-based, and anchor-based methods according to the algorithm frameworks. Segmentation-based methods (e.g.,~\cite{parashar2017scnn,abualsaud2021laneaf}) treat lane detection as a segmentation task but usually require complicated clustering post-processing. Parameter-based methods (e.g.,~\cite{tabelini2021polylanenet}) adopt parameter regression where the parameters are utilized to model the ground-truth lanes. Such methods can run efficiently but often suffer from unsatisfactory accuracy. Anchor-based methods (e.g.,~\cite{qu2021focus,wang2022keypoint}) inherit the idea from generic object detection by formulating lane detection as keypoint detection plus a simple linking operation. Lanes can be detected based on prior anchors and instance-level outputs can be generated directly without complex post-processing.

\begin{figure}[t]
    \centering
    \begin{minipage}{0.40\linewidth}
        \centering
        \begin{subfigure}{1\linewidth}
        \includegraphics[width=1\textwidth]{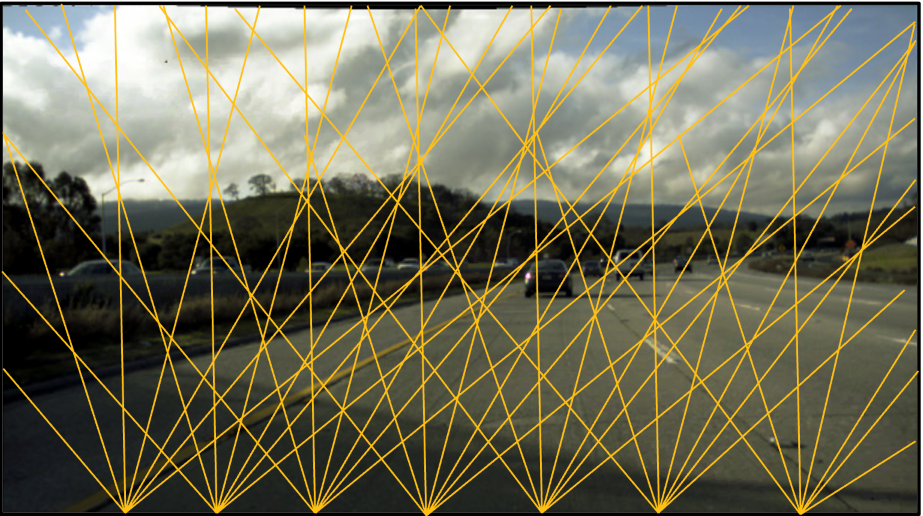}
        \caption{Dense anchors with boundary starting \cite{8624563, tabelini2021cvpr}}
        \end{subfigure}
    \end{minipage} 
    \begin{minipage}{0.40\linewidth}
        \centering
        \begin{subfigure}{1\linewidth}
        \includegraphics[width=1\textwidth]{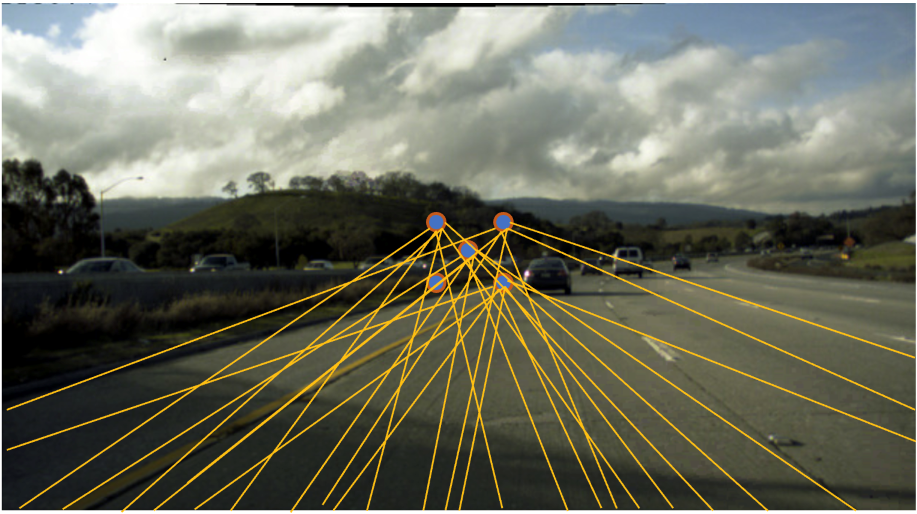}
        \caption{Dense anchors with vanish points starting\cite{su2021structure}}
        \end{subfigure}
    \end{minipage}

    \begin{minipage}{0.40\linewidth}
        \centering
        \begin{subfigure}{1\linewidth}
        \includegraphics[width=1\textwidth]{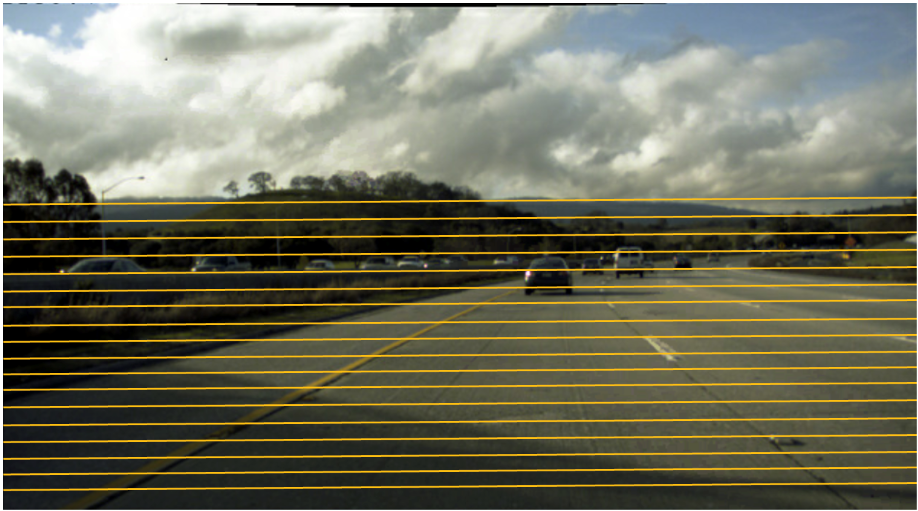}
        \caption{Dense row anchors\cite{qin2020ultra}}
        \end{subfigure}
    \end{minipage} 
    \begin{minipage}{0.40\linewidth}
        \centering
        \begin{subfigure}{1\linewidth}
        \includegraphics[width=1\textwidth]{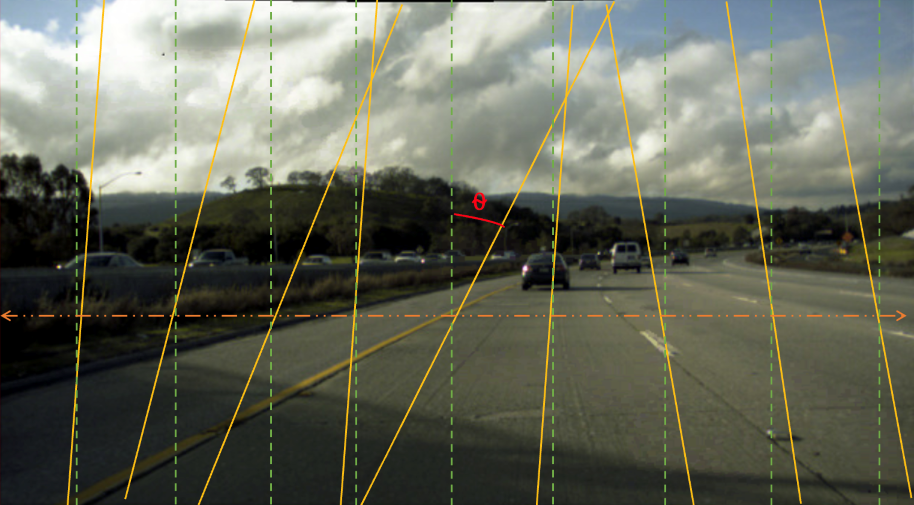}
        \caption{Our sparse anchors}
        \end{subfigure}
    \end{minipage} 

    \caption{Comparison with the anchor setting of state-of-the-art methods. Our method has sparse anchors. In (d), base vertical lane angles (marked with green dash lines) rotate with the predicted angles around defined rotation points to generate dynamic anchors (marked with yellow lines) for every input image.}
     \vspace{-8mm}
    \label{fig:AnchorsComparison}
\end{figure}

We follow the anchor-based paradigm for lane detection in this paper. 
Previous methods usually design hundreds or thousands of anchors but there are only a few lanes in an image of a typical scene. For example,~\cite{tabelini2021cvpr,8624563} design dense anchors based on statistics of the training dataset for lane detection (Figure \ref{fig:AnchorsComparison}a).~\cite{su2021structure} proposes a vanishing point guided anchoring mechanism (Figure~\ref{fig:AnchorsComparison}b) and~\cite{qin2020ultra} adopts dense row anchors (Figure~\ref{fig:AnchorsComparison}c). The limitations of these methods lie in three aspects. First, dense anchors are redundant and will introduce more computation costs. Second, anchors highly depend on the training dataset and require prior knowledge to be designed, which will weaken the transfer ability to different scenes. Third, anchors are fixed and not adaptive for different input images. These analyses naturally motivate us to think about: is it possible to design a lane detector with sparse anchors? Recently, 
~\cite{carion2020end,zhu2020deformable} utilize transformers to formulate object detection as a set prediction problem. Attention mechanism can establish the relationships between sparse object queries and global image context to yield reasonable predictions. We are thus inspired to formulate lane detection into a set prediction problem and only design sparse anchors in the transformer decoder framework.

To this end, we define position-aware sparse lane queries and angle queries (typically 20) instead of traditional dense anchors, as shown in Figure~\ref{fig:AnchorsComparison}d. Each group of lane query and angle query can yield a lane anchor in the transformer decoder. 
The resulting anchors are dynamic and can be adaptive to each specific image. We also employ a two-stage transformer decoder to interact with the proposed sparse queries and refine the lane predictions. Our method, named Sparse Laneformer, is easy-to-implement and end-to-end trainable. Experiments validate the effectiveness of our sparse anchor design and demonstrate competitive performance with the state-of-the-art methods.

Our main contributions can be summarized as follows. (1) 
We propose a simple yet effective transformer-based lane detection algorithm with sparse anchor mechanism. (2) We propose a novel sparse anchor generation scheme where anchors are derived by position-aware lane queries and angle queries. 
(3) We design a two-stage transformer decoder to interact with queries and refine lane predictions.
(4) Our Sparse Laneformer performs favorably against the state-of-the-art methods, e.g., surpassing Laneformer by 2.9\% F1 score and O2SFormer by 0.6\% F1 score with fewer MACs on CULane with the same ResNet34 backbone.


\section{RELATED WORK}

\subsection{Lane Detection}
The lane detection task aims to detect each lane as an instance in a given image. Current deep learning based methods can be mainly divided into three categories:
segmentation-based, parameter-based and anchor-based methods.


\textbf{Segmentation-based Methods.}~Both bottom-up and top-down schemes can be used in segmentation-based methods. 
Bottom-up methods~\cite{abualsaud2021laneaf,qu2021focus,wang2022keypoint,zheng2021resa,xu2020curvelane} tend to find out the active positions of all lanes, and then separate them into different instances with auxiliary features and post-processing.
LaneAF~\cite{abualsaud2021laneaf} predicts a heat map for each keypoint and part affinity fields are used to associate body parts with individuals in the image. FOLOLane~\cite{qu2021focus} produces a pixel-wise heatmap and obtains points on lanes via semantic segmentation. Then a post-processing process is developed to constrain the geometric relationship of adjacent rows and associate keypoints belonging to the same lane instance.
CondLaneNet~\cite{Liu_2021_ICCV} introduces a conditional lane detection strategy to obtain the starting point and then generates the parameters for each lane. 

\textbf{Parameter-based Methods.}~PolyLaneNet~\cite{tabelini2021polylanenet} and LSTR~\cite{liu2021end} consider lanes as polynomials curves, and generate polynomial parameters directly. Parameter-based methods aim to obtain an abstract object and are sensitive to the polynomial parameters (especially the high-order coefficients). Such methods can run efficiently but often suffer from unsatisfactory performance.

\textbf{Anchor-based Methods.}~Anchor-based methods can detect lanes directly without the complex clustering process. The challenge is how to define the lane format and how to design anchors. UFLD~\cite{qin2020ultra} proposes row anchors with dense cells for lane detection. Line-CNN~\cite{8624563} designs the lane as a line anchor, i.e., a ray emitted from the edge of the image, to replace the traditional bounding box.  LaneATT~\cite{tabelini2021cvpr} follows the same line anchor mechanism~\cite{8624563} and adopts anchor-based attention to enhance the representation capability of features.
SGnet~\cite{su2021structure} introduces a vanishing point guided anchoring mechanism to incorporate more priors. CLRNet~\cite{zheng2022clrnet} integrates high-level and low-level features to perceive the global context and local location of lane lines. Laneformer~\cite{han2022laneformer} is a recent transformer-based lane detection method that applies row and column self-attention to obtain lane features. O2SFormer~\cite{zhou2023end} presents a novel dynamic anchor-based positional query to explore explicit positional prior with designing an one-to-several label assignment to optimize model. 
In view of the impressive performance and the efficiency of post-processing, we also adopt the anchor-based paradigm for lane detection. Our method differs from the existing anchor-based methods in three aspects. (1) Existing methods (e.g.,~\cite{tabelini2021cvpr,zheng2022clrnet}) rely on hundreds or thousands of anchors to cover the priors of lane shape, while we design sparse anchors to reduce the algorithm complexity with comparable performance. (2) Compared to Laneformer~\cite{han2022laneformer} and O2SFormer~\cite{zhou2023end}  that are also built on the transformer, we present different attention interactions with a different anchor generation mechanism, and implement a simpler pipeline without the need of auxiliary detectors. 

\begin{figure}[t!]
    \centering
     \vspace{2mm}
    \includegraphics[width=0.8\linewidth]{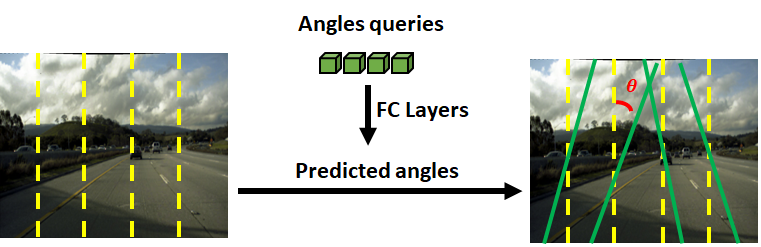}
    \caption{Dynamic anchor generation. The initialized vertical anchors are rotated around the rotation point by $\theta$, which is predicted by angle queries.}
    \label{fig:dynamicanchors}
    \vspace{-6mm}
\end{figure}

\subsection{Generic Object Detection with Sparse Anchors}
Recently, sparse anchor based object detection methods have shown promising performance in a simpler pipeline. For example, DETR~\cite{carion2020end} utilizes sparse object queries that interact with global features via cross attention. The queries are eventually decoded into a series of predicted objects with a set-to-set loss between predictions and ground truths. Sparse R-CNN~\cite{sun2021sparse} also applies set prediction on a sparse set of learnable object
proposals, and can directly classify and regress each proposal for final predictions. Our method is inspired by generic object detection with sparse anchors but has non-trivial designs for lane detection. First, replacing the object queries of DETR directly with lane anchors does not work effectively. We assume that lane detection requires strong priors but queries of DETR can not integrate such priors explicitly. We thus propose position-aware pair-wise lane queries and angle queries to generate anchors. Second, a lane line is represented by equally-spaced 2D points in our method, which is different from the bounding-box format for a generic object. That leads to different attention modules and different transformer decoder designs. 
Third, our anchors are dynamic and adaptive to each specific image during inference, which can decrease the dependencies on the training dataset for anchor design and increase the flexibility of our method.

\section{METHOD}
\subsection{Sparse Anchor Design}

We use equally-spaced 2D points along the vertical direction to represent a lane line. In particular, a lane $\mathcal{P}$ can be represented by a sequence of points, i.e., $\mathcal{P} = \{(x_1, y_1), ..., (x_N, y_N)\}$. The $y$ coordinates are sampled equally along the vertical direction of the image, i.e., $y_i = H_{img}/(N - 1) \times i$, where $H_{img}$ is the height of image and $N$ is the number of sampling points. 

Unlike existing methods (e.g., \cite{tabelini2021cvpr, su2021structure}) that employ extremely dense anchors, we design $K$ sparse anchors ($K=20$ is set as default in our experiments). Each anchor is generated by a pair of learnable lane query and angle query. As shown in Figure~\ref{fig:dynamicanchors}, the $K$ initial anchors are equally sampled along the horizontal direction and each one is a vertical line with $N$ points. We then design a transformer decoder to learn lane queries and angle queries. Each angle query is further fed into a dynamic lane predictor to predict a rotation angle for its associated initial anchor. We also define a rotation point at $y=r\times H$ for each anchor\footnote{In the image coordinate system, the origin is at the top-left corner.} ($r=0.6$ is set as default in our experiments), where $H$ is the height of feature map fed into the transformer decoder. Thus, the predicted anchor $\mathcal{A}$ can be obtained by rotating the initial vertical anchor around this rotation point with the predicted angle. In parallel, lane queries are also fed into the dynamic lane predictor to predict offsets $\mathcal{O}$. The final lane can be obtained by simply combining the predicted anchors and offsets ($\mathcal{P} = \mathcal{A} + \mathcal{O}$).

Our sparse anchor design brings two benefits over previous anchor-based methods. First, existing methods require dense anchors (e.g., 3000 anchors for Line-CNN \cite{8624563} and 1000 anchors for LaneATT \cite{tabelini2021cvpr}), while our anchors are sparse (typically 20) and still can yield comparable lane prediction results. Second, existing methods \cite{8624563, tabelini2021cvpr} depend on the statistics of the training dataset to define anchors and keep anchors unchanged during inference, while our anchors are dynamic since the rotation angle is predicted for each specific image. That helps increase the generality and flexibility of our method.

\begin{figure*}[t!]
    \centering
    \includegraphics[width=0.75\textwidth]{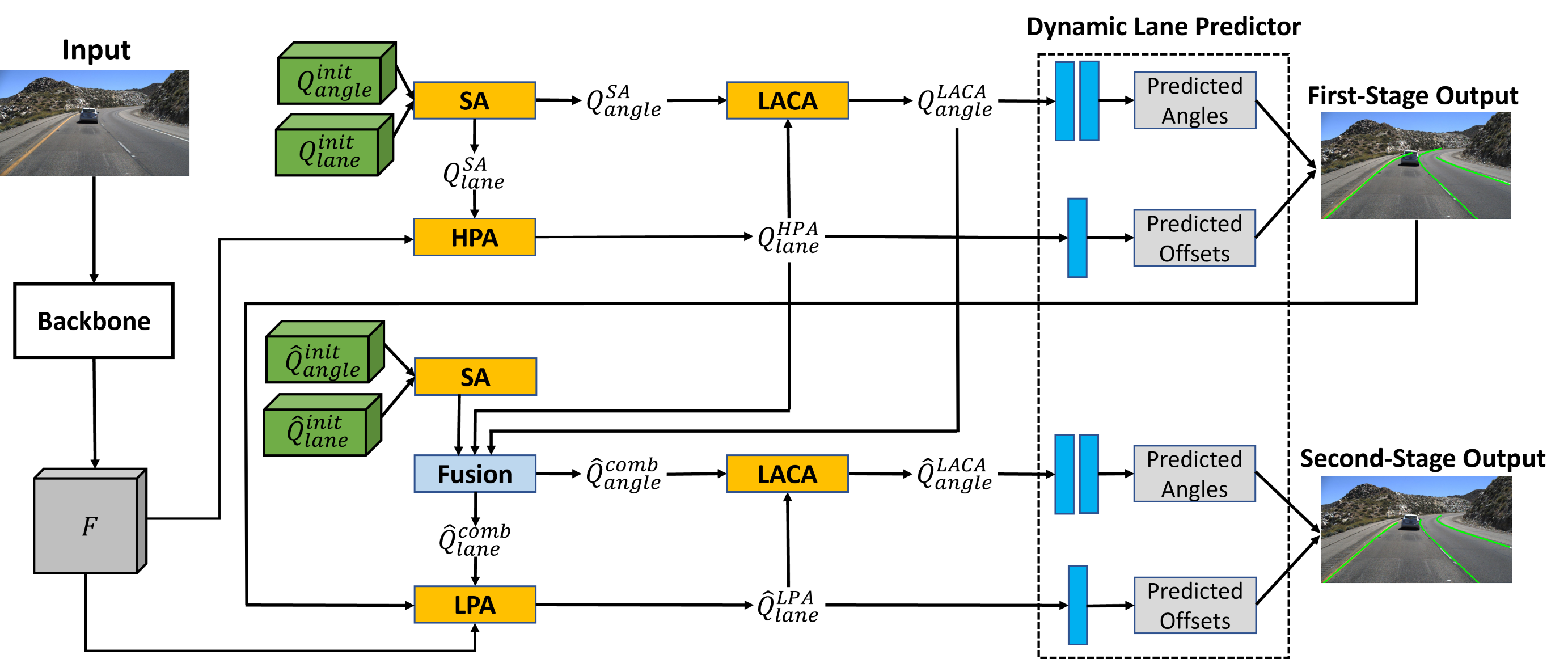}
    \caption{Overview of the proposed Sparse Laneformer. "Fusion" represents Eq. \ref{eq:comb}. See text for details.}
    \label{fig:overall}
    \vspace{-4mm}
\end{figure*}

\subsection{Transformer Decoder Design} 
Our method first adopts a CNN-based backbone to extract feature maps from the input image, then feeds the extracted feature maps into the transformer decoder to generate lane predictions. Specifically, we design a two-stage transformer decoder to perform interaction between features and queries. In the first stage, the initial queries are learned via attention and coarse lanes are predicted via a dynamic lane predictor. In the second stage, the queries and coarse lanes are further refined to output final predictions. Figure \ref{fig:overall} illustrates the overview of the proposed transformer decoder design.

\subsubsection{First-Stage Decoder} 

We first adopt a CNN-based backbone (e.g., ResNet34) to extract feature maps $F \in \mathbb{R}^{C \times H \times W}$ from the input image. In the first-stage decoder, we initialize two types of queries: lane queries $Q^{init}_{lane} \in \mathbb{R}^{C \times H \times K}$  and angle queries  $Q^{init}_{angle} \in \mathbb{R}^{C \times 1 \times K}$ , where the $C$ is the embedding dimension, $H$ is the height of the feature map fed into the decoder, $K$ is the number of anchors that is identical to the number of lane/angle queries. Each lane query has a dimension of $C\times H$ and each angle query has a dimension of $C\times 1$. A pair of lane query and angle query is responsible for generating an anchor. We perform self-attention for the initial lane queries to obtain $Q^{SA}_{lane} \in \mathbb{R}^{C \times H \times K}$, which establishes relationships between different lane queries. Similarly, $Q^{SA}_{angle} \in \mathbb{R}^{C \times 1 \times K}$ is obtained via self-attention to establish relationships between different angle queries.

\begin{figure}
    \centering
    \begin{minipage}{0.80\linewidth}
        \centering
        \begin{subfigure}{1\linewidth}
        \includegraphics[width=1\textwidth]{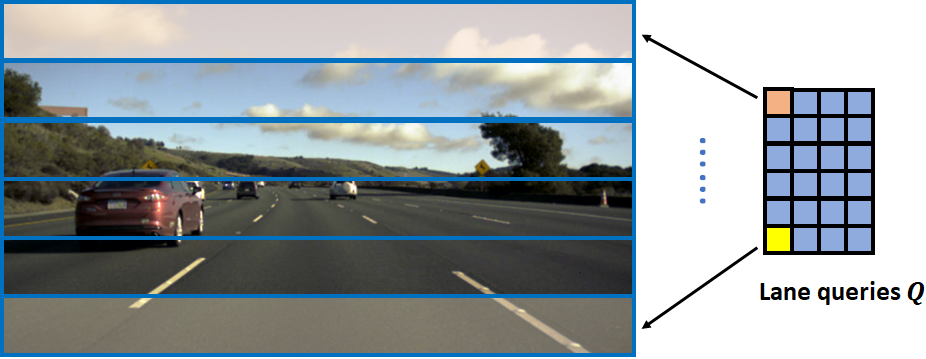}
        \end{subfigure}
    \end{minipage} 

    \caption{Sketch map of Horizontal Perceptual Attention (HPA). Each element in lane queries will only focus on features in one horizontal region of the image.}
    \vspace{-8mm}
    \label{fig:module-HPA}
\end{figure}

\textbf{Horizontal Perceptual Attention (HPA)}. The straightforward way to compute cross attention between features and lane queries is taking $Q^{SA}_{lane}$ as $Q$ and taking the feature map $F$ as $K$ and $V$ in the attention module. That leads to the computation complexity of $O(HWC^{2} + KH^{2}WC)$. We thus propose Horizontal Perceptual Attention (HPA) to improve efficiency. Specifically, as shown in Figure~\ref{fig:module-HPA}, we constrain each row of lane queries only computing cross attention with the corresponding row in the feature map. In other words, each row of lane queries only interacts with a horizontal region context instead of the entire image area. HPA leads to the reduced computation complexity of $O(HWC^{2} + KHWC)$ with reasonable predictions. The output lane queries $Q^{HPA}_{lane} \in \mathbb{R}^{C \times H \times K}$ by HPA can be formulated as:

\begin{equation}
\begin{gathered}
        \tilde{q}^{i}_{lane} = \mbox{MHA}(Q=q^{i}_{lane}, K = f^{i}, V = f^{i})\\
    Q^{HPA}_{lane} = \mbox{Concat}(\tilde{q}^{1}_{lane}, ..., \tilde{q}^{H}_{lane})
\end{gathered}
\end{equation}
where $\mbox{MHA}(\cdot)$ is the multi-head attention~\cite{vaswani2017attention}, $q^{i}_{lane} \in \mathbb{R}^{C \times 1 \times K}$ is the $i$-th row of $Q^{SA}_{lane}$, $f^{i} \in \mathbb{R}^{C \times 1 \times W}$ is the $i$-th row of $F$.

\textbf{Lane-Angle Cross Attention (LACA)}. We further propose Lane-Angle Cross Attention (LACA) to perform cross attention between lane queries and angle queries. Specifically, we take each angle query as $Q$, and each column of 
lane queries as $K$ and $V$. The output angle queries can be formulated as: 
\begin{equation}
\begin{gathered}
    \tilde{q}^{j}_{angle} = \mbox{MHA}(Q=q^{j}_{angle}, K = q^{j}_{lane}, V = q^{j}_{lane}) \\
    Q_{angle}^{LACA} = \mbox{Concat}(\tilde{q}^{1}_{angle}, ..., \tilde{q}^{K}_{angle})
\end{gathered}
\end{equation}
where $\mbox{MHA}(\cdot)$ is the multi-head attention~\cite{vaswani2017attention}, $q^{j}_{angle} \in \mathbb{R}^{C \times 1 \times 1}$ is $j$-th angle query of $Q^{SA}_{angle}$, and $q^{j}_{lane}\in \mathbb{R}^{C \times H \times 1}$is the $j$-th column of $Q^{SA}_{lane}$. Our LACA naturally enables interactions between each angle query and its corresponding lane query.

\textbf{Dynamic Lane Predictor}. We construct a dynamic lane predictor to learn the rotation angle from angle queries and learn offsets from lane queries. Then the initial vertical anchor can be rotated with the predicted angle to generate the dynamic anchor, and the lane can be obtained by combining the dynamic anchor and the predicted offsets. Specifically, the angle queries $Q^{LACA}_{angle} \in \mathbb{R}^{C \times 1 \times K}$ are passed through two fully connected layers and a sigmoid layer to obtain $K$ angles. 
We constrain the predicted angle to be in $(-\pi/3,\pi/3)$ with a simple linear transformation. 
The lane queries $Q^{HPA}_{lane} \in \mathbb{R}^{C \times H \times K}$ are fed to one fully connected layer to output offsets $\mathcal{O} \in \mathbb{R}^{N \times K}$. 
By combining the generated dynamic anchors $\mathcal{A}$ and offsets $\mathcal{O}$, the final $K$ predicted lane lines with $N$ coordinates are obtained $\mathcal{P} = \mathcal{A} + \mathcal{O}$.



\subsubsection{Second-Stage Decoder} 
Based on the coarse lane lines predicted from the first-stage decoder, we build another decoder stage to further correlate lane queries and angle queries and derive more accurate lane line predictions. Similar to the first stage, we initialize lane queries $\hat{Q}^{init}_{lane}$ and angle queries $\hat{Q}^{init}_{angle}$ and perform self-attention to obtain $\hat{Q}^{SA}_{lane}$ and $\hat{Q}^{SA}_{angle}$. Then they are combined with the queries output from the first stage by element-wise addition to obtain $\hat{Q}^{comb}_{lane} \in \mathbb{R}^{C \times H \times K}$ and $\hat{Q}^{comb}_{angle} \in \mathbb{R}^{C \times 1 \times K}$ respectively.
\begin{equation}
\begin{gathered}
 \hat{Q}^{comb}_{lane} = \hat{Q}^{SA}_{lane} + Q^{HPA}_{lane} \\
 \hat{Q}^{comb}_{angle} = \hat{Q}^{SA}_{angle} + Q^{LACA}_{lane}
 \end{gathered}
 \label{eq:comb}
\end{equation}

\textbf{Lane Perceptual Attention (LPA)}. To better optimize the representation of lane queries, we propose Lane Perceptual Attention (LPA) to perform interactions between $\hat{Q}^{comb}_{lane}$ and lane prediction results from the first decoder. 
Specifically, the first-stage transformer decoder outputs $K$ lane predictions, each with $N$ coordinates in the original image. We denote the lane prediction results from the first stage as $T \in \mathbb{R}^{N\times K}$. For each query $(i,j)$ in $\hat{Q}_{lane}^{comb}$, we find its corresponding point in the feature map $F$ via a simple coordinate mapping:
\begin{equation}
    r(i, j) = (i, T(i\times \frac{N}{H}, j))
\end{equation}
All the mapping points for $\hat{Q}_{lane}^{comb}$ can be represented as $R=\{r(i,j)\}$ where $i=1,\dots,H$ and $j=1,\dots,K$. Motivated by~\cite{zhu2020deformable, chen2022persformer}, we learn a fixed number of offsets based on $r(i,j)$ to generate adjacent correlation points. $r(i,j)$ and its correlation points are defined as reference points. These reference points are dynamically adjusted as the network trains, and can enhance the feature representation of lane queries by learning the contextual information of local area.


Our LPA can be formulated as:
\begin{equation} \footnotesize
    \hat{Q}^{LPA}_{lane} = \mbox{DeformAttn}(Q=\hat{Q}^{comb}_{lane}, K=F(R), V=F(R))
\end{equation}
where $\mbox{DeformAttn}(\cdot)$ is the deformable attention \cite{zhu2020deformable}, $\hat{Q}^{comb}_{lane}$ is used as $Q$, the gathered feature $F(R)$ is used as $K$ and $V$. 
The proposed LPA can help enhance the detail recovery capability of lane lines, and also can reduce the computational consumption compared to vanilla cross attention. 

Same as the first stage, angle queries $\hat{Q}^{comb}_{angle}$ also adopt LACA to interact with lane queries $\hat{Q}^{LPA}_{lane} \in \mathbb{R}^{C \times H \times K}$ to generate $\hat{Q}^{LACA}_{angle} \in \mathbb{R}^{C \times 1 \times K}$. $\hat{Q}^{LPA}_{lane}$ and $\hat{Q}^{LACA}_{angle}$ are also fed into the dynamic lane predictor to predict the offsets and angles, respectively. The final lane predictions will be obtained from the second-stage decoder. 


\begin{figure}
    \centering
    \begin{minipage}{0.80\linewidth}
        \centering
        \begin{subfigure}{1\linewidth}
        \includegraphics[width=1\textwidth]{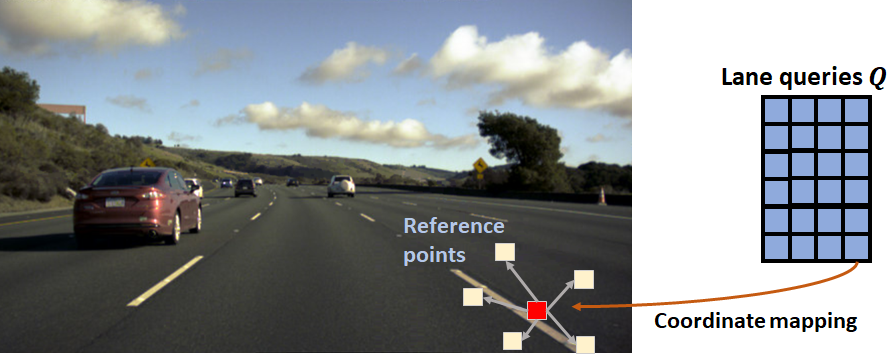}
        \end{subfigure}
    \end{minipage} 
    \caption{Sketch map of Lane Perceptual Attention (LPA). Lane queries only interact with the reference points according to the lane prediction results from the first-stage decoder, which boosts the network to focus on the local contextual details of lane features.}
    \label{fig:module-LPA}
    \vspace{-6mm}
\end{figure}


\subsection{End-to-End Training}

\textbf{Label Assignment.} During training, each ground truth lane is assigned to one predicted lane dynamically as positive samples. We calculate the cost between each predicted lane line and the ground truths. The cost function is defined as:
\begin{equation} 
\begin{aligned}
\mathcal{C}_{assign} &= \mathcal{C}_{reg}+ \mathcal{C}_{cls}  \\
\end{aligned}
\end{equation}
where $\mathcal{C}_{\text {reg }}$ is the regression cost between predicted lanes and ground truth. $\mathcal{C}_{\text {cls }}$ is the cross entropy loss between prediction and ground truth. We then use Hungarian matching for the assignment based on the cost matrix. In this way, each ground truth lane line is assigned to only one predicted lane line.

\textbf{Loss Function.} The loss function includes classification loss,  regression loss and line IoU loss. Classification loss is calculated on all the predictions and labels including positive and negative samples. Regression loss and line IoU loss are calculated only on the assigned samples. The overall loss function is defined as:
\begin{equation}
    \mathcal{L}_{t o t a l}=w_{c l s} \mathcal{L}_{c l s}+w_{reg} \mathcal{L}_{reg}+w_{L I o U} \mathcal{L}_{L I o U}
\end{equation}
where $\mathcal{L}_{c l s}$ is the Focal loss between predictions and labels. $\mathcal{L}_{reg}$ is the L1 regression loss. Similar to~\cite{zheng2022clrnet}, we use line IoU loss to assist in the calculation of regression values. 

\begin{table}[t!] \small
 \begin{center}
\vspace{2mm}
 \begin{tabular}{l|c|c|c}
   \toprule
    Methods &   Backbone  &  F1(\%) & MACs(G)  \\
   \midrule
  PointLane~\cite{chenpoint2019} & ResNet-34 &70.20 &- \\
  CurveLane-S~\cite{xu2020curvelane} & Searched & 71.40 &9.0 \\ 
  CurveLane-M~\cite{xu2020curvelane} & Searched & 73.50 &33.7 \\ 
  CurveLane-L~\cite{xu2020curvelane} & Searched & 74.80 &86.5 \\ 
  ERFNet-HESA~\cite{lee2021robust} & ERFNet &74.20 & -  \\ 
  LaneATT~\cite{tabelini2021cvpr} & ResNet-18 & 75.13 &9.3 \\ 
  LaneATT~\cite{tabelini2021cvpr} & ResNet-34 & 76.68 & 18.0 \\ 
  Laneformer~\cite{han2022laneformer} & ResNet-18 &71.71 &13.8 \\ 
  Laneformer~\cite{han2022laneformer} & ResNet-34 &74.70& 23.0 \\ 
  Laneformer~\cite{han2022laneformer} & ResNet-50 &77.06& 26.2 \\ 
  O2SFormer~\cite{zhou2023end}       &ResNet-18 &76.07 &15.2 \\
  O2SFormer~\cite{zhou2023end}       &ResNet-34 &77.03 &25.1 \\
  O2SFormer~\cite{zhou2023end}       &ResNet-50 &77.83 &27.5 \\
  
   \midrule
   \textbf{Sparse Laneformer} & ResNet-18 & 76.55 &8.5 \\
 \textbf{Sparse Laneformer} & ResNet-34 &77.77& 18.1 \\ 
 \textbf{Sparse Laneformer} & ResNet-50 &77.83  &20.3  \\
   \bottomrule
 \end{tabular}
 \end{center}
   \caption{Lane detection performance comparisons on the CULane dataset.
   }
   \label{table:sotaCULane}
   \vspace{-6mm}
\end{table}

\section{EXPERIMENTS}
\subsection{Experimental Setting}
\textbf{Datasets.}
We conduct experiments on three widely used lane detection datasets: 
CULane~\cite{pan2018spatial},
TuSimple~\cite{tusimple2017} and 
LLAMAS~\cite{behrendt2019unsupervised}.
CULane~\cite{pan2018spatial} is a widely used large-scale dataset for lane detection. It encompasses numerous challenging scenarios, such as crowded roads. The dataset consists of 88.9K training images, 9.7K images in the validation set, and 34.7K images for testing. The images have a size of 1640×590.
TuSimple~\cite{tusimple2017} is one of the most widely used datasets containing only highway scenes. There are 3268 images for training, 358 for validation, and 2782 for testing in this dataset. The image resolution is $1280 \times 720$. 
 LLAMAS~\cite{behrendt2019unsupervised} is a large-scale lane detection dataset containing over 100k images. This dataset relies on the generation from high-definition maps rather than manual annotations. Since the labels of the test set are not publicly available, we used the validation set to evaluate the performance of the algorithm. 

\begin{table}[t!] \small
  \begin{center}
  \vspace{2mm}
  \begin{tabular}{l | c | c | c }
    \toprule
     Methods &   Backbone  & F1 (\%)  & Acc (\%)  \\
    \midrule
        UFLD~\cite{qin2020ultra} & ResNet-34 & 88.02 &95.86 \\
    PolyLaneNet~\cite{tabelini2021polylanenet} & EfficientNet-B0 &90.62& 93.36 \\
    FastDraw~\cite{philion2019fastdraw} & - &93.92 &95.20 \\
    EL-GAN~\cite{ghafoorian2018gan} & - &96.26 &94.90 \\
    ENet-SAD~\cite{hou2019learning} & - &95.92 &96.64 \\ 

    LaneAF~\cite{abualsaud2021laneaf} & DLA-34 & 96.49 & 95.62 \\
    FOLOLane~\cite{qu2021focus} & ERFNet & 96.59& 96.92 \\ 
    CondLane~\cite{liu2021condlanenet} & ResNet-18 &97.01 &95.48  \\  
   CondLane~\cite{liu2021condlanenet} & ResNet-34 & 96.98& 95.37  \\  
    SCNN~\cite{pan2018spatial} & VGG-16 &95.97 &96.53 \\
    
    RESA~\cite{zheng2021resa} & ResNet-34 & 96.93 &96.82 \\

    LaneATT~\cite{tabelini2021cvpr}  & ResNet-18&96.71 &95.57 \\ 
    LaneATT~\cite{tabelini2021cvpr}  & ResNet-34&96.77& 95.63 \\ 

   Laneformer~\cite{han2022laneformer} & ResNet-18 & 96.6 & 96.54  \\
   Laneformer~\cite{han2022laneformer} & ResNet-34 & 95.6 & 96.56 \\
   Laneformer~\cite{han2022laneformer} & ResNet-50 & 96.17 & 96.80 \\

    \midrule
     
    \textbf{Sparse Laneformer} & ResNet-18 & 96.64 & 95.57 \\
    \textbf{Sparse Laneformer} & ResNet-34 & 96.81 & 95.69 \\
    \textbf{Sparse Laneformer} & DLA-34 & 96.72 & 95.36  \\
    \bottomrule
  \end{tabular}
  \end{center}
    \caption{Detection performance comparisons on TuSimple. 
    }
    \label{table:sotaTusimple}
    \vspace{-6mm}
\end{table}

\textbf{Implementation Details.}
During the training and inference process, images from the TuSimple and LLAMAS datasets are resized to $640 \times 360$, while images from the CULane dataset are resized to $800 \times 320$. 
We use random affine transformations (translation, rotation, and scaling) and horizontal flip to augment the images. We set the number of sample points $N$ to 72 and the predefined number of sparse anchors $K$ to 20. In the training process, we use the AdamW optimizer and use an initialized learning rate of 0.003 and a cosine decay learning rate strategy. 
The number of training epochs is set to 300 for TuSimple, 45 for LLAMAS, and 100 for CULane.
During the training of the network, error weights in label assignment $w_{\text {reg }}$ and $w_{\text {cls }}$ are set to 10 and 1, respectively. In particular, the weights in the loss function $w_{c l s}$, $w_{reg}$ and $w_{L I o U}$ are set to 10, 0.5 and 5. For each point in lane queries, $M=25$ reference points are generated in LPA.

\textbf{Evaluation Metrics.}
Following the LaneATT~\cite{tabelini2021cvpr},
F1-measure is taken as an evaluation metric for TuSimple, LLAMAS and CULane, and we also take accuracy(Acc) besides F1 as metrics for the TuSimple dataset.


\subsection{Comparison with State-of-the-Art Methods}

\textbf{Results on CULane.}
Table \ref{table:sotaCULane} shows the comparison results of our method with other state-of-the-art lane detection methods on the CULane dataset. 
Compared to the LaneFormer, our method improves the F1 score by 4.84\%, 3.07\% and 0.77\% on ResNet18 and ResNet-34 and ResNet-50, respectively.
Compared to the latest O2SFormer, our method improves the F1 score by 0.48\% and 0.74\% on ResNet-18 and ResNet-34, respectively. For ResNet-50, the F1 score is same between the two methods. Additionally, our method achieves significant reduction in MACs compared to O2SFormer.
Overall, our method achieves high accuracy and lower computational complexity by leveraging sparse anchors (e.g., 20). These results indicate the competitiveness of our approach in lane detection tasks, surpassing other advanced methods in certain configurations.

\textbf{Results on TuSimple.} Table~\ref{table:sotaTusimple} shows the results of our method and other state-of-the-art lane line detection methods on Tusimple. 
Our Sparse Laneformer achieves 96.81\% F1, 95.69\% Acc with ResNet-34 backbone, which is comparable with the previous SOTA methods. Compared to the most related anchor-based methods~\cite{philion2019fastdraw,tabelini2021cvpr,han2022laneformer}, our method outperforms FastDraw~\cite{philion2019fastdraw} by a large margin. Compared to LaneATT which adopts hundreds of anchors, we can obtain comparable accuracy (slightly worse with ResNet-18 and slightly better with ResNet-34) with only 20 anchors.  Compared to Laneformer, our results are comparable in terms of accuracy, while Laneformer introduces an extra detection process where Faster-RCNN is applied to detect predefined categories of objects firstly (e.g., cars, pedestrians). 


\begin{table}[t!] \small
  \begin{center}
  \vspace{2mm}
  \begin{tabular}{l | c | c|c}
    \toprule
     Methods &   Backbone  & F1 (\%) &MACs(G) \\
    \midrule
    PolyLaneNet~\cite{tabelini2021polylanenet} & EfficientNet-B0 &90.20 &-\\
    LaneATT~\cite{tabelini2021cvpr}  & ResNet-18&94.64 &9.3\\
    LaneATT~\cite{tabelini2021cvpr}  & ResNet-34&94.96 &18.0\\
    LaneATT~\cite{tabelini2021cvpr}  & ResNet-122&95.17 &70.5\\
    LaneAF~\cite{abualsaud2021laneaf}  & DLA-34&96.97 &22.2\\
    \midrule
  
    \textbf{Sparse Laneformer} & ResNet-18 & 96.12 &8.5\\
    \textbf{Sparse Laneformer} & ResNet-34 & 96.56 &17.2\\
    \textbf{Sparse Laneformer} & DLA-34 & 96.32 &15.1\\
    \bottomrule
  \end{tabular}
  \end{center}
    \caption{Detection performance comparisons on LLAMAS. 
    }
    \label{table:sotaLLAMAS}
\vspace{-6mm}

\end{table}

\textbf{Results on LLAMAS.} Table \ref{table:sotaLLAMAS} shows the results of our method and other state-of-the-art lane line detection methods on LLAMAS. Our method can achieve 96.56\% F1 score with ResNet-34, surpassing the most related anchor-based LaneATT by about 1.6\% F1. In addition, our method outperforms the parameter-based PolyLaneNet with 6.3\% F1, and underperforms the segmentation-based LaneAF with 0.6\% F1. However, LaneAF requires complicated clustering post-processing while our method is simpler and more efficient.

\subsection{Ablation study}
 We conduct ablation studies on anchor setting, sparse anchors and two-stage decoder on Tusimple dataset.

\textbf{Effect of Anchor Setting.}
We test different positions of anchor rotation points and different numbers of anchors to analyze the impact of anchor setting on ResNet-34 backbone. The rotation point is at $y=r\times H$ in the image coordinate system (i.e., the origin is the top-left corner). We test the ratio $r$ in a range from 0.5 to 1.0. Table~\ref{table:ablahanchor} shows that our method is not sensitive to the position of anchor rotation point and the number of anchors.
Even with 10 anchors only, our method can reach competitive performance, which validates the effectiveness of our sparse anchor design. We set the position of the rotation point to 0.6 and set the number of anchors to 20 as default in our experiments. 

\textbf{Effect of Sparse Anchors.}
We consider two experiments to analyze the impact of sparse vs. dense anchors. (1) We use 1000 anchors (same amount with LaneATT) and achieve 97.01\% F1 score on TuSimple. From 20 to 1000 anchors, the accuracy increases slightly from 96.81\% to 97.01\% and MACs increase from 17.2G to 19.7G. (2) We select top 20 among the 1000 prior anchors of the original LaneATT and its accuracy decreases largely from 96.77\% to 75.65\%. The results show that our sparse anchor design is non-trivial and our method is not simply built upon existing dense anchor method by reducing anchor amount.

\begin{table}[t] \small
  \begin{center}
  \vspace{2mm}
  \begin{tabular}{c | c | c} 
    \toprule
      position of rotation point  & \# of anchors & F1(\%) \\
         \midrule
     0.5 & 20 & 96.76   \\
     \textbf{0.6} & \textbf{20}& \textbf{96.81}   \\
   0.7 & 20& 96.69  \\
     0.8 & 20 & 96.56  \\
     0.9 & 20 & 96.53  \\
     1.0 & 20 & 96.32  \\
    \midrule
    0.6& 10 & 96.78 \\
     \textbf{0.6} & \textbf{20}& \textbf{96.81}\\
   0.6 & 30& 96.24  \\
    0.6 &40& 96.55\\

    \bottomrule
  \end{tabular}
  \end{center}
    \caption{Ablation study on different positions of rotation points and different numbers of anchors.}
    \label{table:ablahanchor}
    \vspace{-6mm}
\end{table}

\begin{table}[htb] \small
  \begin{center}
  \begin{tabular}{l | c | c }
    \toprule
     Backbone &   Decoder    & F1 (\%) \\
    \midrule
    ResNet-34  & One-Stage  & 95.70   \\
    ResNet-34 & Two-Stage  & \textbf{96.81}   \\
    \bottomrule
  \end{tabular}
  \end{center}
    \caption{Ablation study on the two-stage decoder.}
    \label{table:ablarefinement}
    \vspace{-2mm}
\end{table}

\textbf{Effect of Two-Stage Decoder.} Our method proposes a two-stage transformer decoder to learn anchors and predict lanes. The second-stage decoder takes the coarse lane predictions as input for further refinement. To verify the effect of our transformer decoder, we design a comparison experiment with or without the second-stage decoder with other settings unchanged. Table \ref{table:ablarefinement} shows that the F1 score is decreased by 1.11\% on TuSimple using a one-stage decoder only with the same ResNet-34 backbone, which demonstrates the necessity of our design.



\section{CONCLUSIONS}


In this paper, we propose a transformer-based lane detection framework based on a sparse anchor mechanism. Instead of dense anchors, we generate sparse anchors with position-aware pair-wise lane queries and angle queries. Moreover, we design a two-stage transformer decoder to learn queries for lane predictions. We propose HPA to aggregate the lane features along the horizontal direction, and propose LACA to perform interactions between lane queries and angle queries. We also adopt LPA based on deformable cross attention to perform interactions between lane queries and the coarse lane prediction results from the first stage. Extensive experiments demonstrate that Sparse Laneformer performs favorably against the state-of-the-art methods. In the future, we will further improve the efficiency and accuracy of Sparse Laneformer and extend it for 3D lane detection.

\bibliographystyle{IEEEtran} 
\bibliography{IEEEabrv}

\end{document}